\documentclass[conference]{IEEEtran}
\IEEEoverridecommandlockouts
\usepackage{cite}
\usepackage{amsmath,amssymb,amsfonts}
\usepackage{algorithmic}
\usepackage{graphicx}
\graphicspath{ {./images/} }
\usepackage{textcomp}
\usepackage{xcolor}
\usepackage{caption}
\usepackage{subcaption}
\usepackage{enumitem}

\def\BibTeX{{\rm B\kern-.05em{\sc i\kern-.025em b}\kern-.08em
		T\kern-.1667em\lower.7ex\hbox{E}\kern-.125emX}}

\usepackage{tikz,xcolor,hyperref}
\definecolor{lime}{HTML}{A6CE39}
\DeclareRobustCommand{\orcidicon}{
	\begin{tikzpicture}
		\draw[lime, fill=lime] (0,0) 
		circle [radius=0.16] 
		node[white] {{\fontfamily{qag}\selectfont \tiny ID}};
		\draw[white, fill=white] (-0.0625,0.095) 
		circle [radius=0.007];
	\end{tikzpicture}
	\hspace{-2mm}
}

\foreach \x in {A,B}{\expandafter\xdef\csname orcid\x\endcsname{\noexpand\href{https://orcid.org/\csname orcidauthor\x\endcsname}
		{\noexpand\orcidicon}}
}

\newcommand\copyrighttext{%
	\footnotesize © 2020 IEEE. Personal use of this material is permitted. Permission from IEEE must be 
	obtained for all other uses, in any current or future media, including 
	reprinting/republishing this material for advertising or promotional purposes, creating new 
	collective works, for resale or redistribution to servers or lists, or reuse of any copyrighted 
	component of this work in other works. \href{dx.doi.org/10.1109/ICDMW51313.2020.00043}{DOI: 10.1109/ICDMW51313.2020.00043} }
\newcommand\copyrightnotice{%
	\begin{tikzpicture}[remember picture,overlay]
		\node[anchor=south,yshift=10pt] at (current page.south) {\fbox{\parbox{\dimexpr\textwidth-\fboxsep-\fboxrule\relax}{\copyrighttext}}};
	\end{tikzpicture}%
}
\begin{document}

	\title{Boosting Algorithms for Delivery Time Prediction in Transportation Logistics}

		\author{\IEEEauthorblockN{Jihed Khiari \orcidA{}}
			\IEEEauthorblockA{\textit{Chair for ITS-Sustainable Transport Logistics 4.0} \\
				\textit{ Johannes Kepler University }\\
				Linz, Austria \\
			jihed.khiari@jku.at}
			\and
			\IEEEauthorblockN{Cristina Olaverri-Monreal \orcidB{}}
		\IEEEauthorblockA{\textit{Chair for ITS-Sustainable Transport Logistics 4.0} \\
		\textit{ Johannes Kepler University}\\
		Linz, Austria \\
		cristina.olaverri-monreal@jku.at }
		}
		
	\maketitle
	\copyrightnotice
	\begin{abstract}
		Travel time is a crucial measure in transportation. Accurate travel time prediction is also fundamental for operation and advanced information systems. A variety of solutions exist for short-term travel time predictions such as solutions that utilize real-time GPS data and optimization methods to track the path of a vehicle. However, reliable long-term predictions remain challenging. We show in this paper the applicability and usefulness of travel time i.e. delivery time prediction for postal services. We investigate several methods such as linear regression models and tree based ensembles such as random forest, bagging, and boosting, that allow to predict delivery time by conducting extensive experiments and considering many usability scenarios. Results reveal that travel time prediction can help mitigate high delays in postal services. We show that some boosting algorithms, such as light gradient boosting and catboost, have a higher performance in terms of accuracy and runtime efficiency than other baselines such as linear regression models, bagging regressor and random forest. 

	\end{abstract}
	
	\begin{IEEEkeywords}
		Travel time prediction, Transportation logistics, Ensemble learning, Boosting
	\end{IEEEkeywords}
	
	\section{Introduction}
	
	With the increase of demand volume and e-commerce activities, postal services face more challenges in order to maintain efficiency and customer retention ~\cite{tochkov2015efficiency, morganti2014impact}.
	On the other side, the increasing demand for transport makes it imperative to provide solutions to reduce its environmental impact.
	
	Traffic analysis provides insights into the scope of the problem~\cite{validi2020environmental}. For example, traveling time and origin destination matrices can be examined to determine if a route is the most appropriate and to improve travel schedules~\cite{gonccalves2014smartphone}.

	Reliably predicting delivery time can be a valuable task for operations as well as for customer information systems. Having good estimates of delivery i.e. travel times can allow to optimize delivery routes as well as assignments. At the same time, showing customers timely and reliable delivery times can help avoiding unsuccessful delivery trials, which would entail further trips from the side of the customer to a pick up location or further trips from the side of the delivery company to return the goods to the sender. 
	
	This approach can also contribute to a decrease in cities’ agglomerations in the last kilometers before the delivery and as a consequence it helps to reduce long-term logistic costs and to limit global temperature increases~\cite{kopica2020automated}. Furthermore, this can result in higher customer trust and satisfaction, which can be lucrative for the postal service company. 
	
	A traditional machine learning approach to travel time prediction is regression analysis. It comprises a large number of techniques to estimate the relationship between a set of predictors (i.e features) and a dependent variable ~\cite{friedman2001elements}. It is defined as follows: 
	\begin{equation}
	\hat{f}: x_i,\theta \rightarrow \mathbb{R} \mbox{ such that } \hat{f}(x,\theta)=f(x_i)=y_i, \forall x_i \in X, y_i \in Y
	\end{equation}
	where $f(x_i)$ denotes the true unknown function which is generating the samples' target variable and $\hat{f}(x_i, \theta)=\hat{y}_i$ is an approximation dependent on the feature vector $x_i$ and an unknown parameter vector $\theta \in \mathbb{R}^n$ (given by a given induction model $M$). Evidently, this approximation will be as good as the adequacy of $M$ to the dependence structure of $f$ as well as the relevancy of the input feature space $X$.\\ 
	
	To address this problem, boosting algorithms have emerged as a relevant solution as they tend to perform well in terms of accuracy and runtime 
	~\cite{nielsen2016tree, taieb2014gradient}. 
	
	The idea behind these algorithms is to be able to generate strong learners with low bias by combining weak learners~\cite{schapire1990strength}. 
	
	Boosting algorithms work as follows:

	\begin{enumerate}
		\item Sequentially learn an ensemble of weak models such as decision trees or stumps, such that each model tries to correct the mistakes made by the previous one.
		\item Combine these into one strong model, primarily reducing bias and also variance to achieve high prediction power.
		\item Can in principle be used to improve any supervised learning, e.g. classification or regression algorithm.
	\end{enumerate}

	Various boosting-based approaches were proposed in the last years~\cite{drucker1997improving}. These approaches are commonly used for machine learning competitions thanks to their generalization power~\cite{abou2018xgboost}. 
	
	Therefore, we rely on them to answer the following research questions:
	\begin{enumerate}
		\item How does the training period and retraining frequency affect performance?
		\item Given the variety of available boosting algorithms, which one is better suited for this task?
		\item Considering possible deployment, which setup (retraining frequency, prediction target, model) would offer higher usability and consistent results?
	\end{enumerate}
	
	Our main contribution consists in conducting extensive experiments to address these questions and demonstrate the potential of applying machine learning techniques to temporal data in an industrial environment. \\
	
	In the remainder of this paper, we first present relevant work that has been carried out for a similar problem. Then, we detail the methodology and the experimental setup. Results are presented and discussed before introducing possible future directions and concluding the paper.
	
	\section{Related Work}
	Numerous studies have been carried out to devise and propose solutions to travel time prediction for intelligent transportation systems. Such solutions cover not only transportation of goods~\cite{brandstatter2020efficient}, but also public~\cite{mendes2015improving, hassan2016feature, khiari2018metabags} and private transport~\cite{hoch2015ensemble}. However, most studies focus on short term predictions that are restricted to a relatively small geographical location. In this case, real-time or high frequency GPS data or sensor data is required in order to ensure high accuracy. Long term predictions that involve trips that are typically longer than one hour and can span over several cities remain challenging due to the complexity of the setting and/or the unavailability of real-time sensor data. 
	
	Our work is focused on machine learning approaches to address long-term travel time prediction for logistics. Therefore, we present in this section relevant work that has been done in a similar context.
	
	In~\cite{lin2005review}, the authors review the main approaches used to collect data and estimate travel times, namely statistical approaches, machine learning approaches, and a historical data estimation method. The work in~\cite{miura2010study} proposes kriging; a statistical approach that relies on spatial predictions for car travel time. The authors of~\cite{wu2004travel} rely on support vector regression to predict travel time for highway traffic data. Similarly, the study presented in~\cite{lee2009knowledge} focuses on urban networks and proposes a solution based on meta-rules using data from location-based services. 
	
	On the other hand, many studies, e.g.~\cite{gmira2020travel} and~\cite{liu2018time}, focus on the last mile delivery for goods and services. For example the work described in~\cite{gmira2020travel} relies on a long short term memory network~\cite{hochreiter1997long} for predicting travel times related to home delivery of large appliances such as a television or furniture. Whereas,~\cite{liu2018time} presents a food delivery use case and proposes a solution that combines optimization heuristics and machine learning approaches including linear models, e.g., the least absolute shrinkage and selection operator (LASSO) and ridge regression, and nonlinear models, e.g., support vector regression and random forest.

	Furthermore, a different type of algorithms can be applied such as nonparametric distribution-free regression models as described and applied to floating car data from a logistics company in~\cite{simroth2010travel} 
	
	Boosting algorithms have been developed and improved during the last years (e.g. Boost-by-Majority~\cite{freund2001adaptive}, AdaBoost~\cite{freund1999short}, LogitBoost~\cite{cai2006using}. In ~\cite{zhang2015gradient}, gradient boosting is suggested for travel time prediction applied to highway sensor data. In addition, the work in ~\cite{li2019travel} proposes two tree-based ensembles, namely random forest and gradient boosting.

	We contribute to the field of research by investigating different boosting approaches to predict travel times for delivery companies. To the authors' knowledge, there aren't any other works in the literature that tackle this problem in a similar way.

	\section{Methodology}
	
	In this work, we focus on temporal data collected by a postal service company that operates across the country. We have access to chronological trip information from a period of 7 months. With the goal of predicting travel times using a machine learning approach, we devise different scenarios related to retraining, size of training and testing sets, as well as prediction target. For this task, we focused on models that typically have a low runtime without compromising accuracy, namely boosting algorithms. Although machine learning competitions such as Kaggle are often won by ensemble learning methods, boosting algorithms -in particular- have emerged as a common winning solution to many of the competitions ~\cite{nielsen2016tree, taieb2014gradient}. 
	
	We therefore considered six variants of boosting algorithms; gradient boosting~\cite{friedman2002stochastic}, Adaboost~\cite{freund1999short}, extreme gradient boosting~\cite{chen2015xgboost}, light gradient boosting~\cite{ke2017lightgbm}, histogram gradient boosting~\cite{pedregosa2011scikit}, and Catboost~\cite{prokhorenkova2018catboost}. Additionally, we investigated other baselines such as random forest~\cite{breiman2001random}, tree-based bagging regressor~\cite{breiman1996bagging}, and linear regression models e.g., the least absolute shrinkage and selection operator (LASSO) and ridge regression~\cite{friedman2001elements}.  
	
	We compare in this work, results from six boosting algorithms that are described under \ref{baselines} in order to investigate which algorithm or sub-group of algorithms is appropriate for the prediction task at hand. For all boosting ensembles in this work, we considered regression decision trees as the boosting algorithms' weak learners.
	
	To do so, we propose different training and testing scenarios, two prediction targets, and we measure performance of these boosting algorithms as well as other baselines in terms of prediction error (mean absolute error (MAE), root mean squared error (RMSE)) and runtime. We describe in this section the sample, the preprocessing steps and the testing scenarios.

	\subsection{Data Sample Description}
	
	Thanks to a research collaboration, we were able to obtain trip information data from the postal service company in Austria.

	The data included stops in several cities and the following additional parameters: trip number, trip description, stop number, client name, address, scheduled delivery time, and actual delivery time. 
	The sample contained trip information from a total of 7 months, stating from March 2019 to September 2019 and January 2020. 

	Table \ref{Table: Dataset} shows some key aspects of the dataset at hand. 
	
	\begin{table}[t!]
		\centering
		\caption{Dataset summary. For all information following the total number of trips, we report the mean value and the standard deviation (std) as follows: mean $\pm$ std}
		\label{Table: Dataset}
		\begin{tabular}{ |l|l| } 
			\hline
			Total Number of Trips & 193864 \\ 
			\hline
			Number of Trips per Day & 811 $\pm$ 480\\ 
			\hline
			Number of Trips per Month & 24233 $\pm$ 2494\\ 
			\hline
			Number of Stops per Trip & 6 $\pm$ 3 \\ 
			\hline
			Number of Cities per Trip & 5 $\pm$ 3  \\ 
			\hline
			Trip Duration & 4.55 $\pm$ 4.19 hours \\
			\hline
			Trip Delay & 0.71 $\pm$ 7.26 hours \\
			\hline		
		\end{tabular}

	\end{table}
	
	We note that the number of trips per day has a relatively high standard deviation. This can be explained by the variance between the number of trips on weekdays compared to that on weekends. Table \ref{tab: nb_trips} shows the mean and standard deviation of the number of trips per type of day: weekday, Saturday, or Sunday.

\begin{figure}[b!]
		\includegraphics[width=1.0\columnwidth]{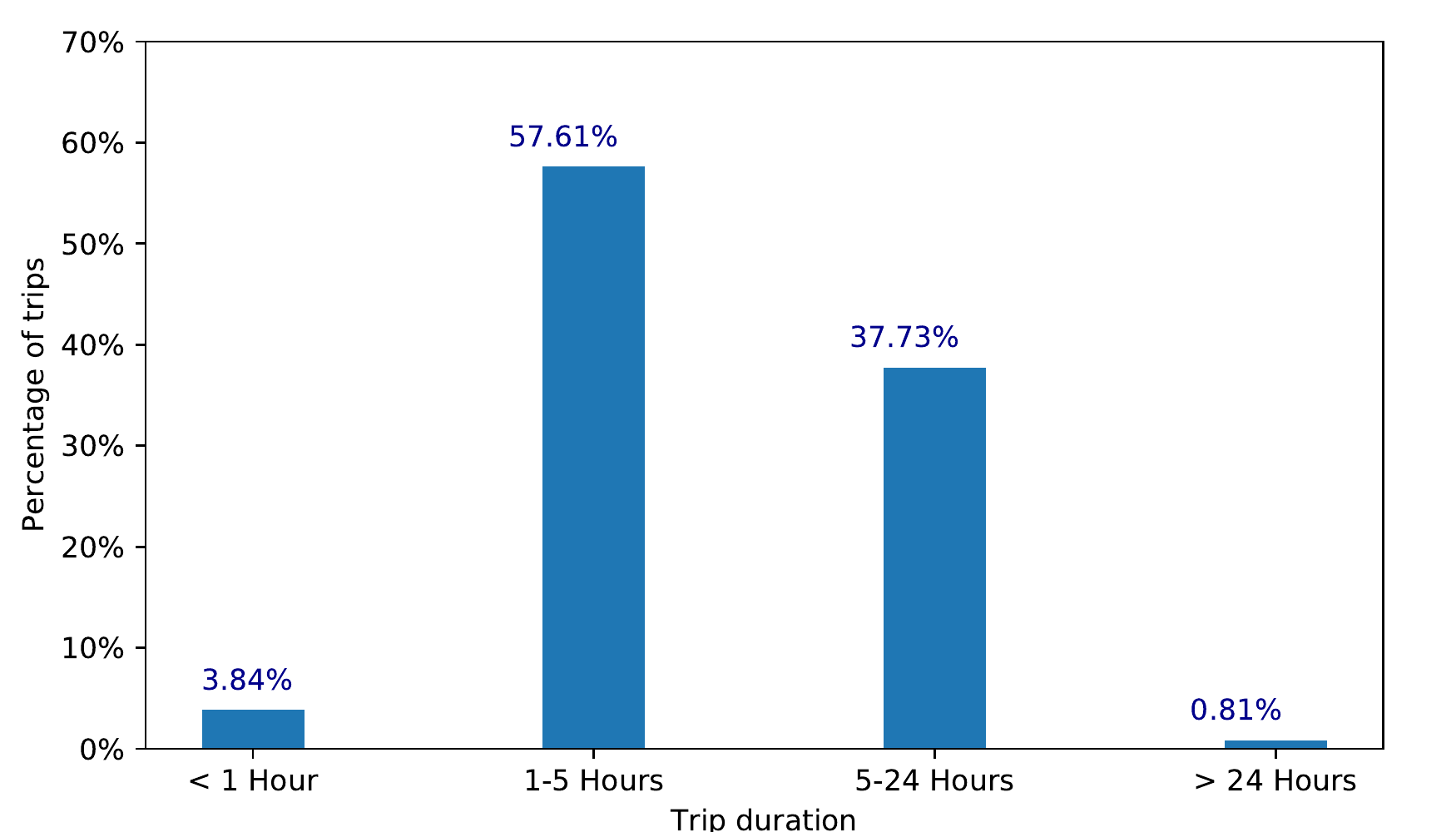}
		\caption{Trips duration distribution}
		\label{fig:duration}
	\end{figure}

	Similarly, trip duration and trip delay are characterized by a high variance due to the variety of trips occurring in the dataset. Figure \ref{fig:duration} 
	illustrate the distribution of trips' duration.
	In Figure \ref{fig:duration}, we can see that the majority of trips take between 1 and 24 hours, whereas a minority of the trips are of less than an hour or require more than 24 hours for completion.

	\subsection{Preprocessing}
	Our focus in this work is on the trip level. We therefore preprocessed the raw dataset in order to characterize the trips. For each trip, we extracted and/or generated the following features from the raw data:
	
	\begin{itemize}
		\item \textbf{Trip id:} a unique identifier for each trip
		\item \textbf{Number of cities:} the number of cities that the trip includes
		\item \textbf{Number of stops:} the number of stops through the trip
		\item \textbf{Temporal features:} month, week number, day of the month, type of day (Monday-Sunday), hour, minute
		\item \textbf{Scheduled trip duration:} the duration of the trip given by the scheduled arrival times in the raw data
	\end{itemize}
	
	We considered two prediction targets for our task: 
	
	\begin{itemize}
		\item \textbf{Trip duration:} It refers to the actual duration of a given trip, meaning the time it takes between the actual arrival at its first stop and the actual arrival at its last stop.
		\item \textbf{Trip delay:} It refers to the cumulated delay that the trip had between its first and last stops. 
	\end{itemize}. 
	
	We conducted two sets of independent experiments in order to investigate whether one target is better suited for this task, i.e. is associated with a higher prediction accuracy.

	\subsection{Considered Testing Scenarios}
	\label{sec:setup}
	To train and test the data, we conducted several experiments under different scenarios that were performed on a computer with the following specifications:
	\begin{enumerate}
		\item Processor: Intel(R) Core(TM) i7 - 10510U CPU@1.8GHz
		\item RAM: 16 GB
		\item GPU: GeForce MX250 \\
	\end{enumerate}
	
	For all scenarios, the testing was performed on the last three months of the dataset, thus allowing a fair comparison of all scenarios in terms of mean average error (MAE), root mean square error (RMSE), and training time. The scenarios are described as follows:
	
	\begin{itemize}
		\item \textbf{Scenario 0:} No retraining; the first 4 months are used for training and the remaining 3 are used for testing.
		\item \textbf{Scenario 1:} Retraining every month; one month is used for testing, 3 months for training.
		\item \textbf{Scenario 2:} Retraining every two weeks; two weeks are used for testing, 6 weeks for training.
		\item \textbf{Scenario 3:} Retraining very week; 1 week is used for testing, 3 weeks for training.
		\item \textbf{Scenario 4:} Retraining every day; 1 day is used for testing, 3 days for training. \\
	\end{itemize}
	
	In scenarios 1-4, we have maintained the same ratio between training and testing set sizes.
	
	\begin{table}[!t]
		\centering
		\caption{Number of trips per day sorted by week or weekend days. Mean value and standard deviation (std) are reported as follows: mean $\pm$ std.}
		\label{tab: nb_trips}
		\begin{tabular}{|l|l|}
			\hline
			Type of Day & Number of Trips \\
			\hline
			Weekday & 1084.57 $\pm$ 237.42 \\
			Saturday & 198.23 $\pm$ 23.54 \\
			Sunday & 48.88 $\pm$ 14.98 \\
			\hline
		\end{tabular}
		
	\end{table}	
	
	\subsection{Baselines}
	\label{baselines}
	To study the performance of boosting methods, we considered the following baseline approaches:
	\begin{itemize}
		\item \textbf{Linear regression (lr)} assumes the regression function is linear in the inputs. The least square approach is often used to fit the linear regression model ~\cite{friedman2001elements}. 
		\item \textbf{Ridge regression (ri)} shrinks the regression coefficients by imposing a penalty on their size ~\cite{hoerl1970ridge}. 
		\item \textbf{Lasso (la)} is a shrinkage method like ridge regression. It is used to estimate sparse coefficients~\cite{tibshirani1996regression}.
		\item \textbf{Decision tree (dt)} is a non-parametric supervised learning method used for classification and regression. The goal is to create a model that predicts the value of a target variable by learning simple decision rules inferred from the data features~\cite{breiman1984classification}.
		\item \textbf{Bagging regression (br)} is an ensemble meta-estimator that fits base regressors each on random subsets of the original dataset and then aggregate their individual predictions to form a final prediction ~\cite{breiman1996bagging}.
		\item \textbf{Random forest (rf)} A random forest is a meta estimator that fits a number of decision tree classifiers on various sub-samples of the dataset and uses averaging to improve the predictive accuracy and control over-fitting ~\cite{breiman2001random}. 	\item \textbf{Gradient boosting (gb)} builds an additive model in a forward stage-wise fashion; it allows for the optimization of arbitrary differentiable loss functions. In each stage a regression tree is fit on the negative gradient of the given loss function ~\cite{friedman2002stochastic}.
		\item \textbf{Adaboost (ab)} fits a sequence of weak learners on repeatedly modified versions of the data. The predictions from all of them are then combined through a weighted majority vote to produce the final prediction ~\cite{freund1999short}.
		\item \textbf{Histogram-based gradient boosting (hgb)} is an experimental implementation of gradient boosting trees inspired by light gradient boosting ~\cite{ke2017lightgbm} introduced by Scikit-learn ~\cite{pedregosa2011scikit}.
		\item \textbf{Extreme gradient boosting (xgb)} provides an efficient implementation of the gradient boosting algorithm with regard to computational efficiency which often yields better model performance ~\cite{chen2015xgboost}.
		\item \textbf{Catboost (cb)} is a gradient boosting implementation put forth by Yandex. It generally shows state-of-the-art results with a reduced training time and higher support for descriptive data formats ~\cite{prokhorenkova2018catboost}.
		\item \textbf{Light gradient boosting (lgb)} is a gradient boosting framework that is designed to be distributed and efficient thus resulting in faster training speed, lower memory usage and better accuracy. Furthermore, lgb supports parallel and GPU learning and is capable of handling large-scale data ~\cite{ke2017lightgbm}. \\     	 
	\end{itemize}

	For the sake of simplicity, the abbreviations between parentheses are used in the legends of the graphical representations in the next section to indicate the corresponding baseline in the results.
	
	In addition, to further investigate the methods' scalability, we devised an experiment where we set the number of samples of the training set and report the time needed to fit the model. The number of samples varies between 1000 samples and 150,000 samples.
	
	\section{Results}
\begin{figure}[t]
		\includegraphics[width=1.0\columnwidth]{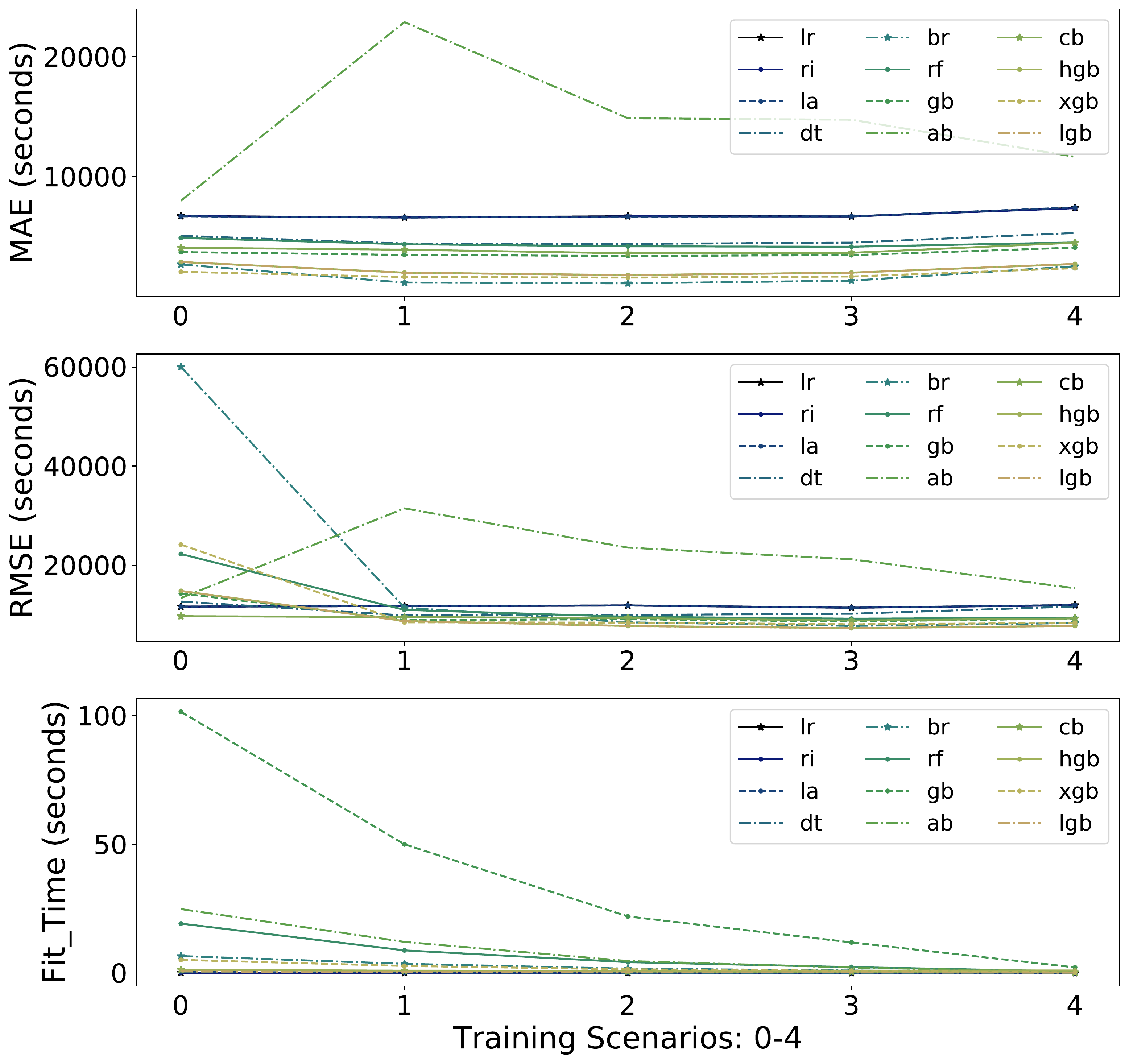}
		\caption{Comprehensive results for trip duration (target 1) prediction: MAE, RMSE, fit time per scenario}
		\label{fig:df1}
	\end{figure}

\begin{figure}
		\includegraphics[width=1.0\columnwidth]{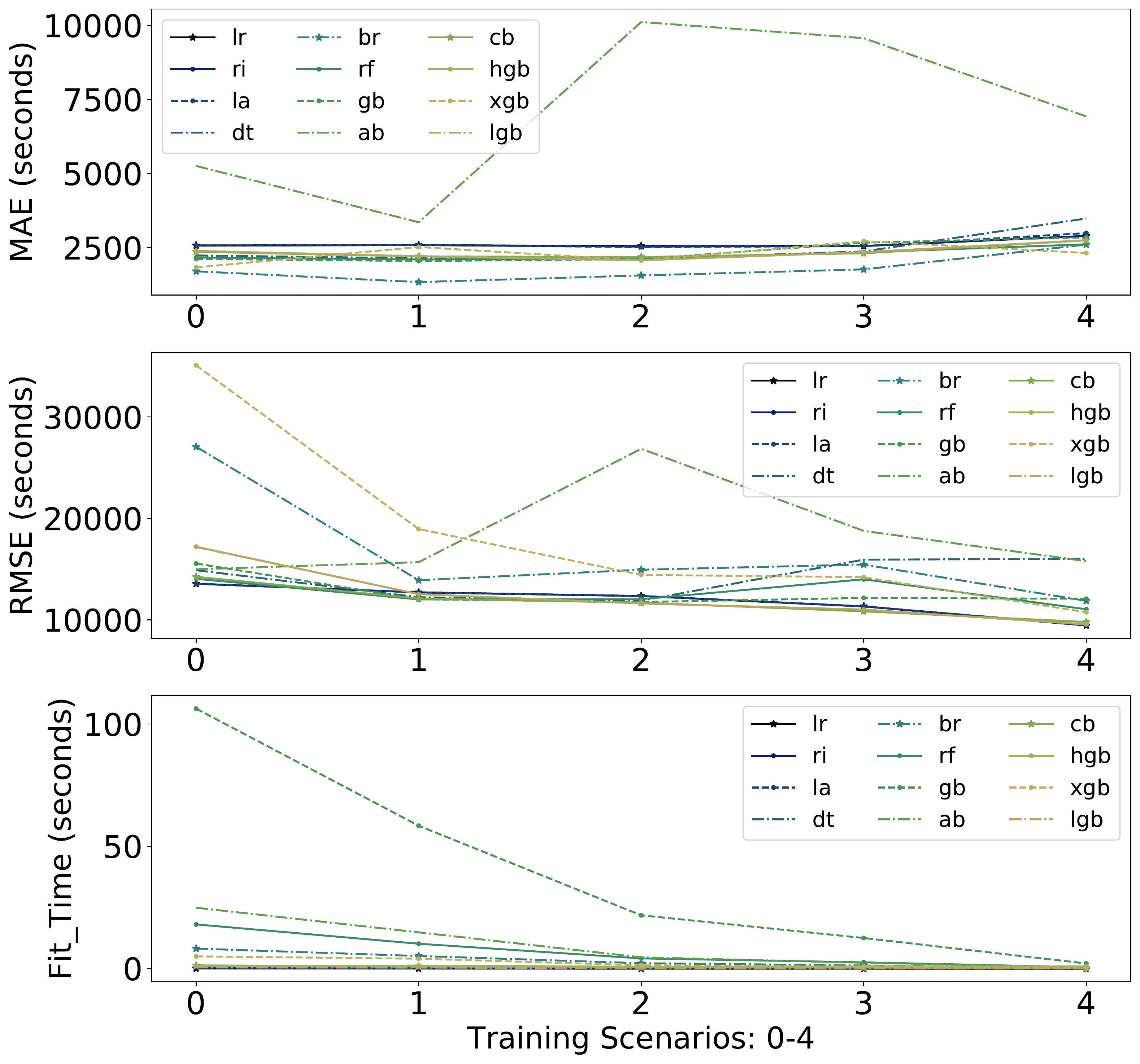}
		\caption{Comprehensive results for trip delay (target 2) prediction: MAE, RMSE, fit time per scenario}
		\label{fig:df2}
	\end{figure}

\begin{figure}
		\includegraphics[width=1.0\columnwidth]{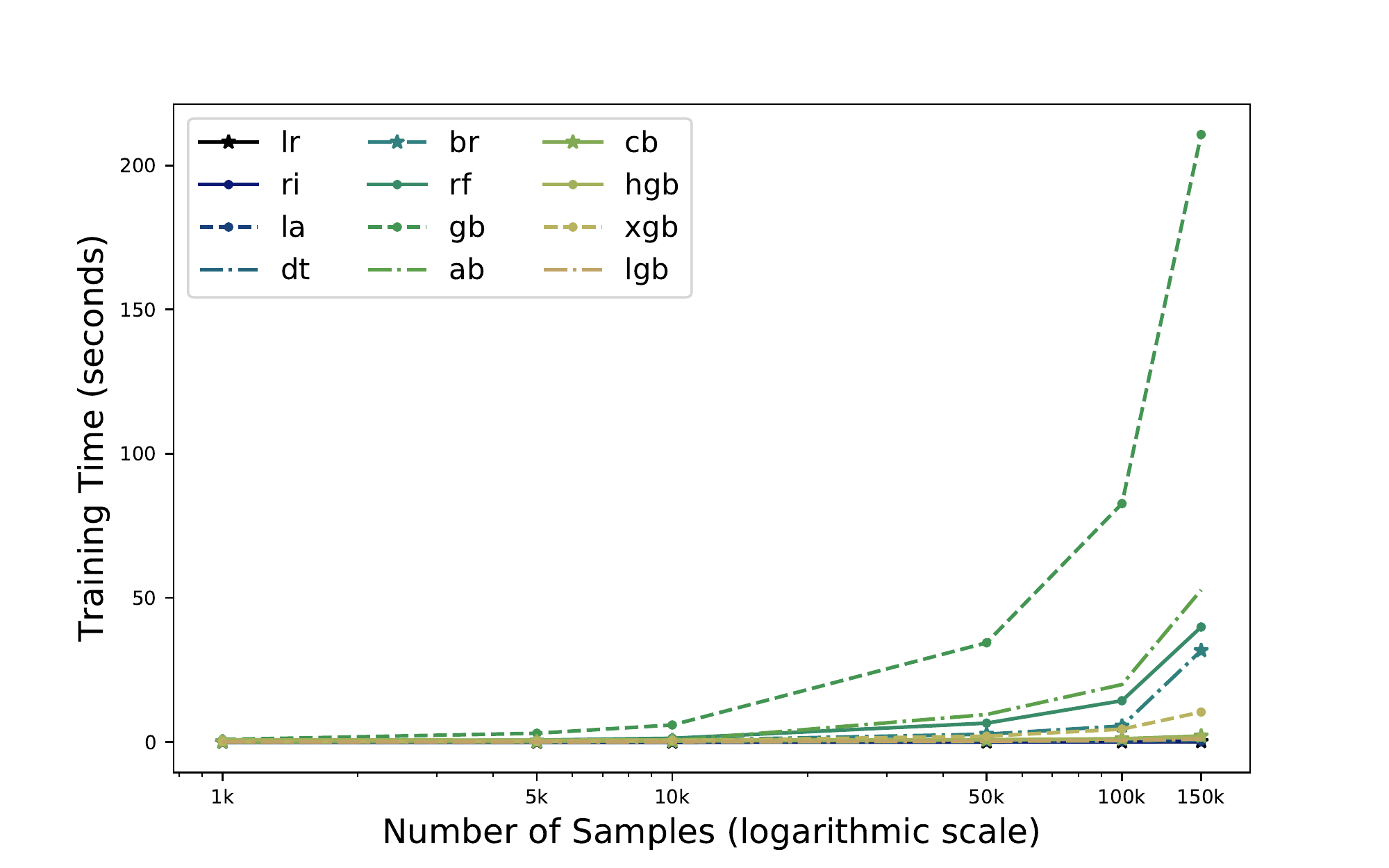}
		\caption{Scalability results: training time per number of samples}
		\label{fig:scale}
	\end{figure}

	We detail in this section the obtained results from the performed analysis. Figure \ref{fig:df1} and Figure \ref{fig:df2} illustrate the results considering the trip duration and the trip delay respectively as the prediction target. We report MAE, RMSE, and fit time per training scenario. In Figure \ref{fig:scale}, we show the results of the scalability experiment.

	MAE measures the average magnitude of errors in a set of predictions.  It is the average over the test sample of the absolute differences between prediction and actual observation where all individual differences have equal weight, as shown in Equation \ref{eq:mae}.
	
	RMSE, on the other hand, is a quadratic scoring rule that also measures the average magnitude of the error. It is the square root of the average of squared differences between prediction and actual observation, as shown in Equation \ref{eq:rmse}.
	While both MAE and RMSE are measures of average model prediction error, they can be interpreted differently. Since the errors are squared before they are averaged, RMSE penalizes large errors more. Therefore, if a model's RMSE in a given scenario is particularly high, it indicates that there are instances where the difference between observations and predictions is large. MAE, however, gives a more intuitive idea about the average error.
	
	We report both MAE and RMSE in order to have a better idea about the errors. \\
	
	\begin{equation}
	MAE = \frac{1}{n}\sum_{i=1}^{n}\left | y_{i} - x_{i} \right |
	\label{eq:mae}
	\end{equation}
	
	\begin{equation}
	RMSE = \sqrt{\frac{1}{n}\sum_{i=1}^{n}(y_{i} - x_{i})^{2}}
	\label{eq:rmse}
	\end{equation}

	In scenarios that include retraining, the reported results are averaged over the testing period. For instance, in the case of Scenario 1, where retraining is done every month and the testing period is of one month, the retraining will take place three times in order to cover the common testing period for all the scenarios i.e. the 3 last months of the dataset. 
	
	The runtimes required to fit a model are reported in Figures \ref{fig:df1} and \ref{fig:df2} as the averaged values of the three runs. 
	
	For each training scenario, all the above-mentioned approaches are trained and tested according to the scenarios defined under Section \ref{sec:setup}.

\begin{figure}
		\includegraphics[width=1.0\columnwidth]{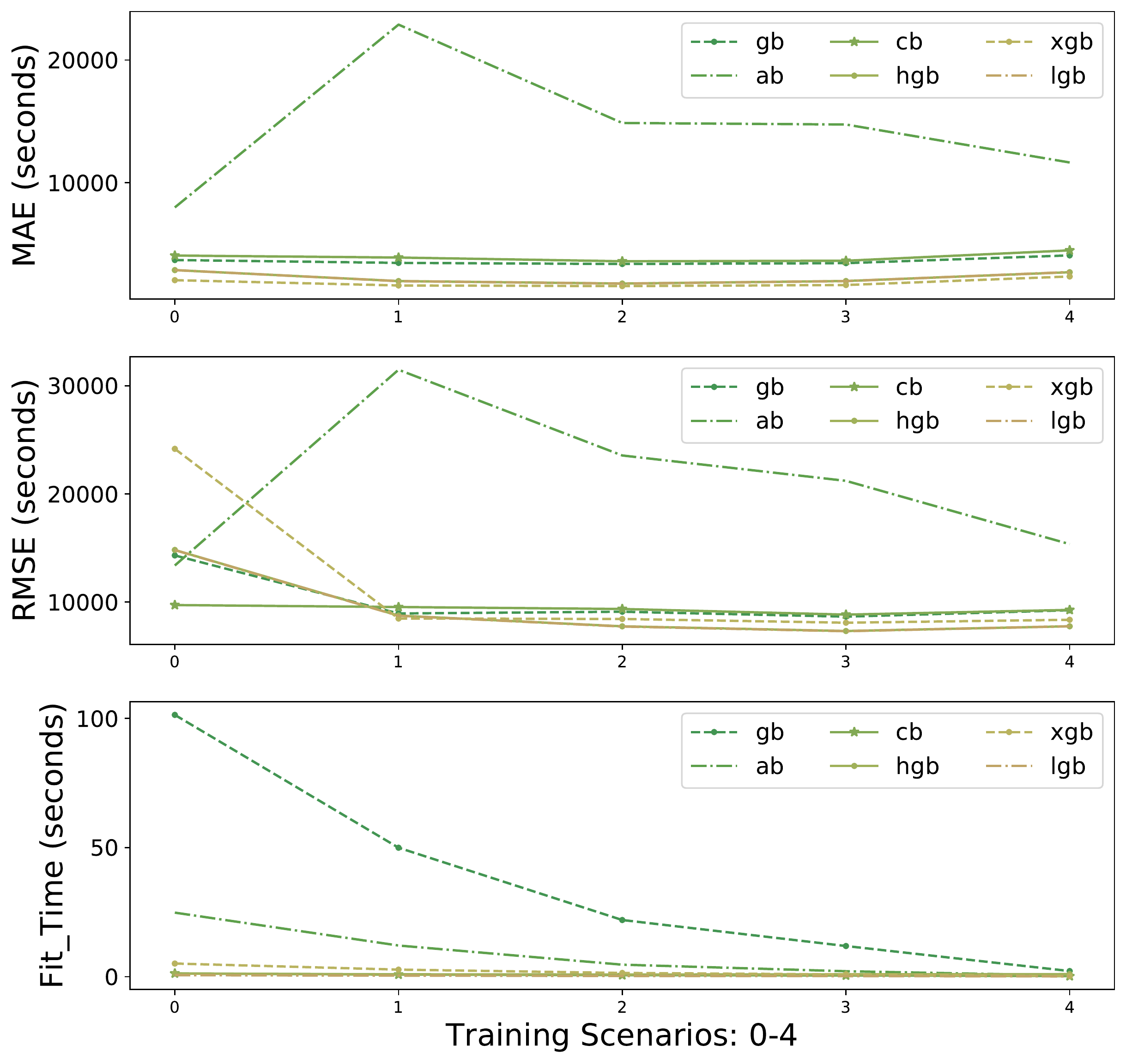}
		\caption{Boosting algorithms results for trip duration (target 1) prediction: MAE, RMSE, fit time per scenario}
		\label{fig:df1_b}
	\end{figure}

\begin{figure}
		\includegraphics[width=1.0\columnwidth]{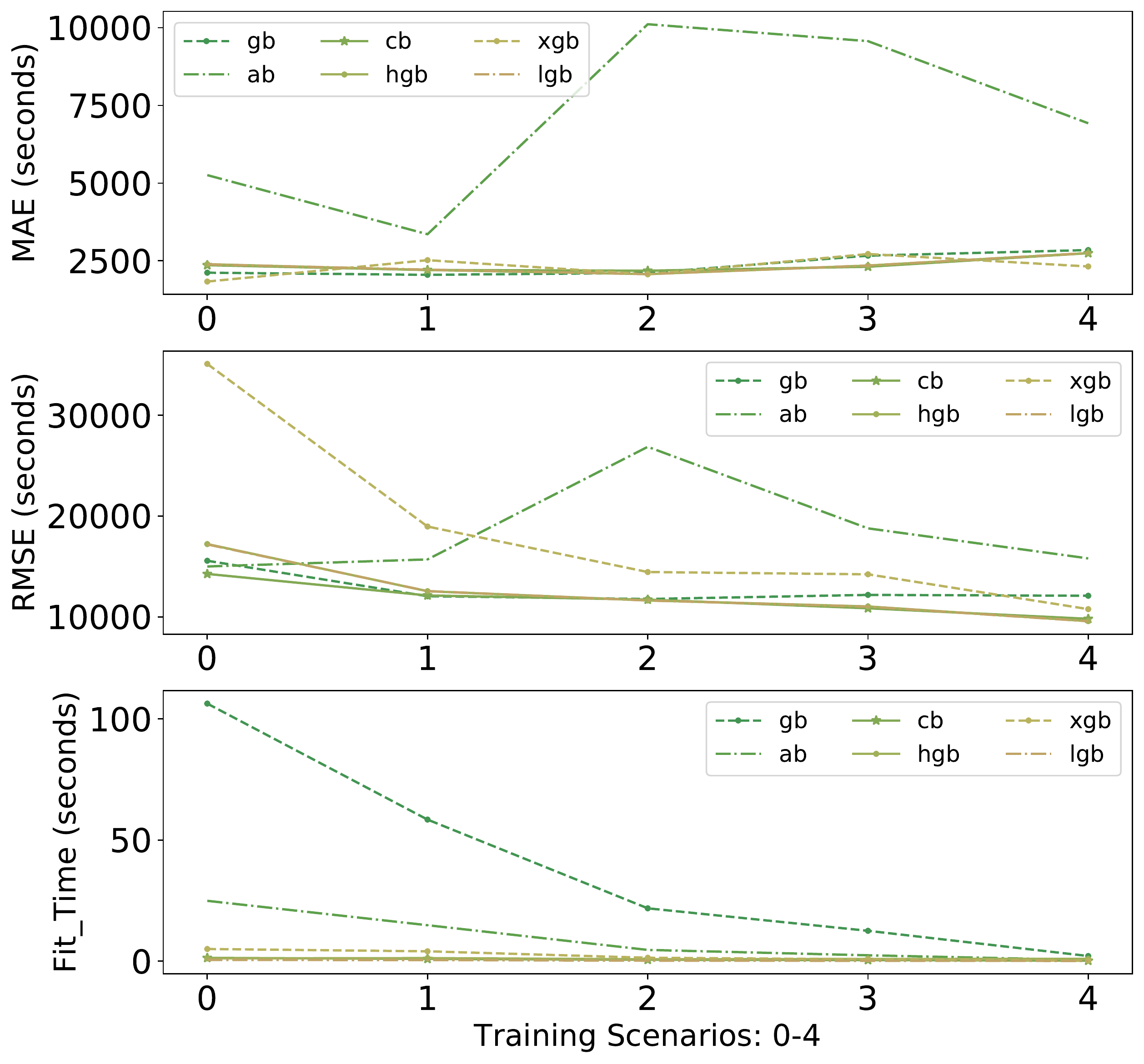}
		\caption{Boosting algorithms results for trip delay (target 2) prediction: MAE, RMSE, fit time per scenario}
		\label{fig:df2_b}
	\end{figure}

\begin{figure}
		\includegraphics[width=\columnwidth]{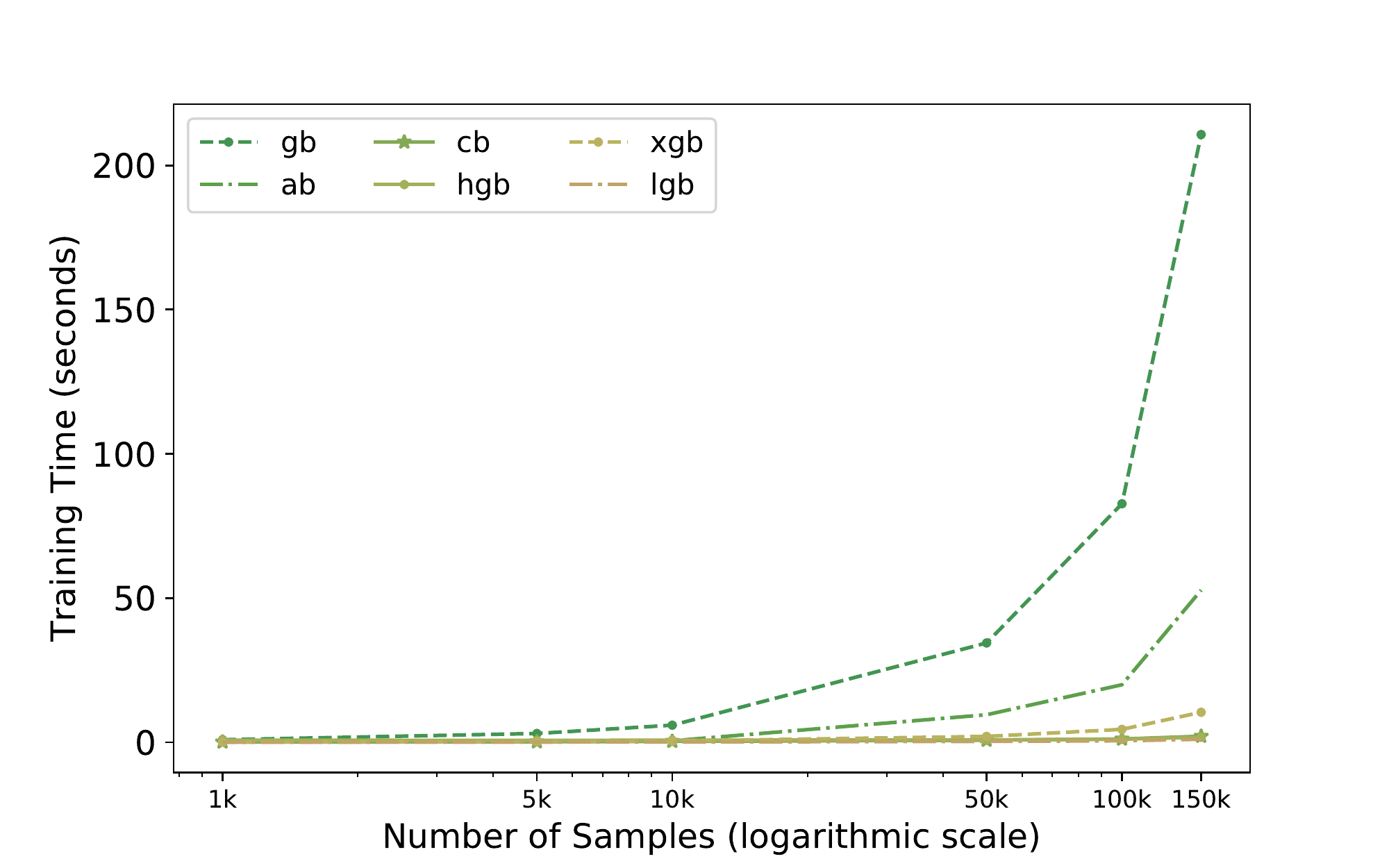}
		\caption{Scalability results for boosting algorithms: training time per number of samples}
		\label{fig:scale_b}
	\end{figure}

	For more clarity, we show in Figures \ref{fig:df1_b}, \ref{fig:df2_b}, and \ref{fig:scale_b} results only for boosting algorithms. These figures share the same characteristics of the previous figures showing comprehensive results.
	Tables~\ref{tab:df1_mae}, \ref{tab:df1_rmse}, \ref{tab:df2_mae}, \ref{tab:df2_rmse}, \ref{tab:df1_fit}, and \ref{tab:df2_fit} show the numerical values illustrated in Figures~\ref{fig:df1_b} and \ref{fig:df2_b}. \\
	
	In order to analyze the results, we consider in the following paragraphs different aspects separately. 
	
	\subsection*{MAE Comparison}
	
	We note that for the trip duration prediction (target 1) results illustrated in Figure \ref{fig:df1}, the MAE for all models -except Adaboost- does not vary considerably across the scenarios. Similarly, the MAE for the trip delay prediction (target 2) results illustrated in \ref{fig:df2} show low variability. The results also show that for both prediction targets, boosting algorithms, the bagging regressor has the lowest MAE in most scenarios.
	
	\subsection*{RMSE Comparison} 
	On the other hand, the RMSE values shown in figures \ref{fig:df1} and \ref{fig:df2} vary across scenarios. For both prediction targets, we note a relatively higher error for scenario 0 (no retraining) followed by lower values in scenarios 1-4. In the case of predicting trip duration, the errors are  rather constant in these scenarios, with light gradient boosting, catboost and extreme gradient boosting showing the lowest errors. As for trip delay prediction, we note more variability across scenarios. While the error decreases in scenarios 1-4 for light gradient boosting, catboost, and lasso, we note an increase for the remaining algorithms either for scenario 2 (bagging regressor, adaboost) or for scenario 3 (decision tree, random forest). We note however that the RMSE values are generally lower when predicting the trip delay than they are when predicting trip duration.
	
	\subsection*{Fit Time Comparison}
	For both prediction targets, fit time decreases for all models between scenarios 0 and 4. This is expected as the training period (from 4 months in scenario 0 to 3 days in scenario 4) and thus the number of training data samples decreases between the scenarios. Furthermore, we note that fit time for catboost, light gradient boosting, and histogram gradient boosting, varies very little and remains the lowest across scenarios. This finding is also confirmed through the scalability experiment results shown in Fig.~\ref{fig:scale}; while we note for most models an increase in run time when the number of samples increases to 150k, the runtime remains almost constantly low for for catboost, light gradient boosting, and histogram gradient boosting. This shows that these models scale well and are able to handle large datasets without requiring more time.  
	
	\subsection*{Comparing scenarios}
	As the training period is reduced from 4 months in scenario 0 to 3 days in scenario 4, we note that the MAE shows a slight increase in scenario 4 suggesting that the smallest training period might not be sufficient to learn the delivery patterns. As for RMSE, we note that the values are considerably higher for scenario 0 suggesting that when considering a long period for training, there can be high errors to which RMSE is sensitive. However, the light gradient boosting, catboost and histogram gradient boosting algorithms tend to have the lowest errors across scenarios, not to mention a low training time.
	
	\subsection*{Comparing boosting algorithms}
	Figures \ref{fig:df1_b}, \ref{fig:df2_b}, and \ref{fig:scale_b} show that different boosting algorithms perform differently. For instance, Adaboost has high errors compared to the remaining boosting algorithms. In terms of RMSE in particular, extreme gradient boosting has high errors. Furthermore, gradient boosting and Adaboost have the highest run times. However, light gradient boosting, catboost and histogram gradient boosting have consistent low errors and a low runtime.

	\begin{table}
		\centering
		\caption{Mean MAE values for trip duration (target 1) prediction per scenario and algorithm. Lowest values in bold.}
		\label{tab:df1_mae}
		\begin{tabular}{|c|c c c c c|}
			\hline
			\textbf{Scenario}&\textbf{0}&\textbf{1}&\textbf{2}&\textbf{3}&\textbf{4} \\ \hline
			
			\textbf{gb}&3705.78&3473.12&3382.91&3453.64&4083.62     \\
			\textbf{ab}&7992.86&22886.91&14872.31&14748.51&11654.34 \\
			\textbf{hgb}&2884.93&1992.71&1790.8&1995.54&2712.61     \\
			\textbf{xgb}&\textbf{2064.44}&\textbf{1634.23}&\textbf{1583.95}&\textbf{1671.04}&\textbf{2366.07}    \\
			\textbf{cb}&4070.97&3909.9&3606.54&3647.89&4493.57      \\
			\textbf{lgb}&2886.11&1985.05&1787.43&1989.44&2727.22    \\
			\hline 
		\end{tabular}
	\end{table}

	\begin{table}
		\centering
		\caption{Mean RMSE values for trip duration (target 1) prediction per scenario and algorithm. Lowest values in bold.}
		\label{tab:df1_rmse}
		\begin{tabular}{|c|c c c c c|}
			
			\hline
			\textbf{Scenario}&\textbf{0}&\textbf{1}&\textbf{2}&\textbf{3}&\textbf{4} \\ \hline

			\textbf{gb}&14316.76&8952.85&9112.93&8658.01&9248.27       \\
			\textbf{ab}&13387.8&31479.54&23556.0&21210.27&15361.73     \\
			\textbf{hgb}&14822.34&8755.87&7764.27&7330.93&7766.87      \\
			\textbf{xgb}&24168.32&\textbf{8483.69}&8442.23&8098.27&8368.24      \\
			\textbf{cb}&\textbf{9730.45}&9547.22&9367.47&8846.92&9281.33        \\
			\textbf{lgb}&14806.02&8742.16&\textbf{7743.63}&\textbf{7312.04}&\textbf{7770.22}      \\
			\hline
			
		\end{tabular}
	\end{table}

	\section{Discussion}
	The results show that some boosting algorithms, namely light gradient boosting, Catboost and histogram gradient boosting are the algothims that scale best in terms of runtime compared to all other considered baselines. In terms of error metrics, they also show consistently low values. Although the bagging regressor and LASSO can show relatively low values for MAE and runtime, they do not perform as well in terms of RMSE. Given the small difference in results for these boosting algorithms across scenarios 1-3, it can be challenging to choose which scenario will be better for deployment. 
	The results demonstrate however that these boosting algorithms can yield promising results under varying scenarios. Retraining every week (Scenario 3) seems to ensure a particularly good tradeoff between accuracy and runtime.

	\section{Conclusion and Future Work}
	
	The results confirm the potential of predicting travel time for postal services. The detailed results show that predicting the delay or the trip duration can help mitigate big delays in delivery, e.g. delays that are higher than 5 hours, by ensuring errors that do not exceed one hour in terms of MAE and 3 hours in terms of RMSE. We note that predicting the delay may yield relatively low errors and that even retraining with a small frequency, e.g. every month, can yield low errors and might be sufficient to improve operation and customer information. Furthermore, relying on light gradient boosting, catboost and histogram gradient boosting guarantees a low running time without compromising accuracy, thus simplifying deployment and usability. As a possible next step, we consider including weather data and intermediate GPS data as further data sources in order to improve the accuracy of delivery time predictions. Further research can also incorporate spatial correlations between different locations as well as a measure of trip similarity to better capture the mobility patterns and give more specific results.
	
	\begin{table}
		\centering
		\caption{Mean MAE values for trip delay prediction (target 2) per scenario and algorithm. Lowest values in bold.}
		\label{tab:df2_mae}
		\begin{tabular}{|c|c c c c c|}
			\hline
			\textbf{Scenario}&\textbf{0}&\textbf{1}&\textbf{2}&\textbf{3}&\textbf{4} \\ \hline
			
			\textbf{gb}&2117.35&2048.65&2109.16&2665.52&2845.53\\   
			\textbf{ab}&5257.4&3355.17&10113.14&9568.47&6926.45\\   
			\textbf{hgb}&2390.55&2204.56&2072.3&2346.01&2746.69\\  
			\textbf{xgb}&\textbf{1832.24}&2521.23&\textbf{2064.82}&2719.01&\textbf{2317.53}\\ 
			\textbf{cb}&2359.64&\textbf{2203.25}&2181.98&\textbf{2308.78}&2745.71\\   
			\textbf{lgb}&2385.23&2214.9&2071.54&2348.18&2751.03\\   
			\hline                                     
		\end{tabular}
	\end{table}

	\begin{table}
		\centering
		\caption{Mean RMSE values for trip delay prediction (target 2) per scenario and algorithm. Lowest values in bold.}
		\label{tab:df2_rmse}
		\begin{tabular}{|c|c c c c c|}
			\hline
			\textbf{Scenario}&\textbf{0}&\textbf{1}&\textbf{2}&\textbf{3}&\textbf{4} \\ \hline
			
			\textbf{gb}&15561.4&12046.71&11769.11&12176.67&12090.51     \\
			\textbf{ab}&14995.14&15693.84&26857.42&18772.93&15795.59    \\
			\textbf{hgb}&17215.71&12543.08&11628.83&11030.67&9578.56    \\
			\textbf{xgb}&35077.34&18955.01&14443.77&14217.43&10765.18   \\
			\textbf{cb}&\textbf{14256.46}&\textbf{12120.74}&11706.43&\textbf{10851.09}&9805.78     \\
			\textbf{lg}b&17181.14&12554.27&\textbf{11620.82}&11010.33&\textbf{9565.39}    \\
			\hline                                    
		\end{tabular}
	\end{table}

\begin{table}[t]
	\centering
	\caption{Mean fit time values (in seconds) for trip duration (target 1) prediction per scenario and algorithm. Lowest values in bold.}
	\label{tab:df1_fit}
	\begin{tabular}{|c|c c c c c|}
		
		\hline
		\textbf{Scenario}&\textbf{0}&\textbf{1}&\textbf{2}&\textbf{3}&\textbf{4} \\ \hline
		
		\textbf{gb}&101.41&49.95&21.95&11.88&2.18   \\
		\textbf{ab}&24.81&12.07&4.65&2.11&0.49      \\
		\textbf{hgb}&1.23&0.92&0.84&0.91&1.0        \\
		\textbf{xgb}&5.09&2.74&1.44&0.88&0.21       \\
		\textbf{cb}&1.25&0.8&0.58&0.46&0.21         \\
		\textbf{lgb}&\textbf{0.59}&\textbf{0.41}&\textbf{0.27}&\textbf{0.22}&\textbf{0.13}       \\
		\hline
		
	\end{tabular}
\end{table}

\begin{table}[t]
	\centering
	\caption{Mean fit time values (in seconds) for trip delay (target 2) prediction per scenario and algorithm. Lowest values in bold.}
	\label{tab:df2_fit}
	\begin{tabular}{|c|c c c c c|}
		\hline
		\textbf{Scenario}&\textbf{0}&\textbf{1}&\textbf{2}&\textbf{3}&\textbf{4} \\ \hline
		
		\textbf{gb}&106.29&58.43&21.83&12.55&2.14      \\
		\textbf{ab}&24.92&14.86&4.66&2.44&0.51         \\
		\textbf{hgb}&1.16&1.28&0.8&0.92&0.97           \\
		\textbf{xgb}&5.04&4.09&1.42&0.8&0.21           \\
		\textbf{cb}&1.36&0.96&0.58&0.47&0.2            \\
		\textbf{lgb}&\textbf{0.54}&\textbf{0.49}&\textbf{0.22}&\textbf{0.17}&\textbf{0.13}          \\
		
		\hline
		
	\end{tabular}
\end{table}

		\section*{Acknowledgment}
		This work was supported by the Austrian Ministry for Climate Action, Environment, Energy, Mobility, Innovation and Technology (BMK) Endowed Professorship for Sustainable Transport Logistics 4.0. We would like to thank the Austrian Post for providing us with a data sample.

	\bibliographystyle{IEEEtran}
	\bibliography{jk_jabref}
	
\end{document}